\documentclass[runningheads]{llncs}

\usepackage{graphicx}
\usepackage{tabularx} 
\usepackage[colorlinks = true,
            linkcolor = blue,
            urlcolor  = blue,
            citecolor = blue,
            anchorcolor = blue]{hyperref}
\newcolumntype{Y}{>{\centering\arraybackslash}X}

\begin{document}
\title{End-to-end Malaria Diagnosis and 3D Cell Rendering with Deep Learning}

\author{\textbf{Vignav Ramesh} \\
	Saratoga High School\\
	Saratoga, CA 95070 \\
	\texttt{rvignav@gmail.com}}
\authorrunning{V. Ramesh}

\maketitle              
\begin{abstract}
Malaria is a parasitic infection that poses a significant burden on global health. It kills one child every 30 seconds and over one million people annually. If diagnosed in a timely manner, however, most people can be effectively treated with antimalarial therapy. Several deaths due to malaria are byproducts of disparities in the social determinants of health; the current gold standard for diagnosing malaria requires microscopes, reagents, and other equipment that most patients of low socioeconomic brackets do not have access to. In this paper, we propose a convolutional neural network (CNN) architecture that allows for rapid automated diagnosis of malaria (achieving a high classification accuracy of 98\%), as well as a deep neural network (DNN) based three-dimensional (3D) modeling algorithm that renders 3D models of parasitic cells in augmented reality (AR). This creates an opportunity to optimize the current workflow for malaria diagnosis and demonstrates potential for deep learning models to improve telemedicine practices and patient health literacy on a global scale. Our website is accessible \href{https://topdocmedicine.wixsite.com/topdoc}{here}.

\end{abstract}
\section{Introduction}
\subsection{Current Malaria Diagnosis Methods}
Malaria is a parasitic infection that poses a significant burden on global health. It kills one child every 30 seconds and over one million people annually [1]. If diagnosed in a timely manner, however, most people can be effectively treated with antimalarial therapy. Several deaths due to malaria are byproducts of disparities in the social determinants of health. Malaria is endemic to developing regions such as Sub-Saharan Africa and Southeast Asia, areas of which have limited access to hospitals and laboratory equipment (microscopes, dyes and indicators, volumetric flasks, and other common lab apparatus) that enable diagnoses. Even in areas with adequate medical facilities, current modalities of diagnosis impose a significant financial burden on healthcare systems. These geographic and socioeconomic barriers prevent patients with malaria from receiving a diagnosis and treatment plan within 24 hours, thereby allowing malaria to progress to fatally severe illness [2].

Excluding clinical methods, malaria is currently diagnosed in two ways: via microscopy and via a rapid diagnosis test (RDT). A microscopy test—currently the gold standard of malaria diagnosis—requires a health professional to identify a malaria parasite in a blood smear on a microscope slide. Not only does this require a bulky, expensive microscope, but to ensure accuracy, the blood smear must be stained (often with the Giemsa stain) to morphologically differentiate the nucleus and cytoplasm of the parasite [3]. This stain requires a reagent (Giemsa powder), methanol, and other agents that most patients do not have access to and may not have the medical background necessary to work with. For an RDT, a drop of blood mixed with a lysing agent is placed adjacent to a buffer in an RDT cassette. The blood and buffer are mixed with a labeled antibody and are flushed along the cassette strip; directly proportional to the blood’s antigen concentration (parasite density), a certain quantity of antigen-antibody complex will accumulate on the test line of the strip [4]. Patients in low socioeconomic brackets without immediate access to healthcare professionals may not be able to obtain the cassettes, lysing agents, buffers, and labeled antibodies that RDTs require.

Moreover, even in areas with access to necessary medical and diagnostic equipment, a shortage of medical professionals prevents patients from receiving necessary diagnoses. There is currently a critical shortage of medical professionals in developing countries; for instance, Africa has 2.3 healthcare workers per 1000 residents, leading to overbooking and inadequate care for many patients [5]. Doctors can provide treatment plans in a shorter time period and to more clients, however, if the process of diagnosing a given patient for malaria is streamlined. To this end, we propose the models described in the following section.

\subsection{Contributions} 
In this paper, we present a convolutional neural network (CNN) architecture that allows for rapid automated diagnosis of malaria, as well as a deep neural network (DNN) based three-dimensional (3D) modeling algorithm that renders 3D models of parasitic cells in augmented reality (AR) and on our website, thereby improving current telemedicine practices (as doctors and patients alike can better visualize and comprehend underlying pathology) while also encouraging patient education and health literacy.

This diagnostic tool functions to: 1) empower patients to actively engage in their own care without having to enter a hospital, and 2) provide patients with the knowledge and resources to reach out to physicians at the right time. This is especially useful for two demographics: people living in rural or under-resourced areas where access to hospitals or microscopes is scarce, and patients who wish to understand their current condition but may be unable or unwilling to visit a hospital due to the coronavirus disease 2019 (COVID-19) pandemic and physician shortage.

\section{Methodology and Results}

\subsection{Data}
To complete this study, we utilized a database comprising 27,588 labeled cell-level images (parasitized and uninfected) from the National Institutes of Health (NIH) [6]. There were three components to our approach. First, we trained our model on the NIH data over 25 epochs. Second, we saved the model weights in Hierarchical Data (H5) format. Third, we used a TensorFlow- and Keras-based pruning algorithm to to eliminate unnecessary values in the weight tensors and thus decrease the weight file’s necessary storage space in order to easily host the algorithm on our web server. The end result is a deployed online tool for rapid malaria diagnoses; users can upload cell-level images into our algorithm and will receive a diagnosis in seconds as shown in Fig. 1.

\begin{figure}
\includegraphics[width=\textwidth]{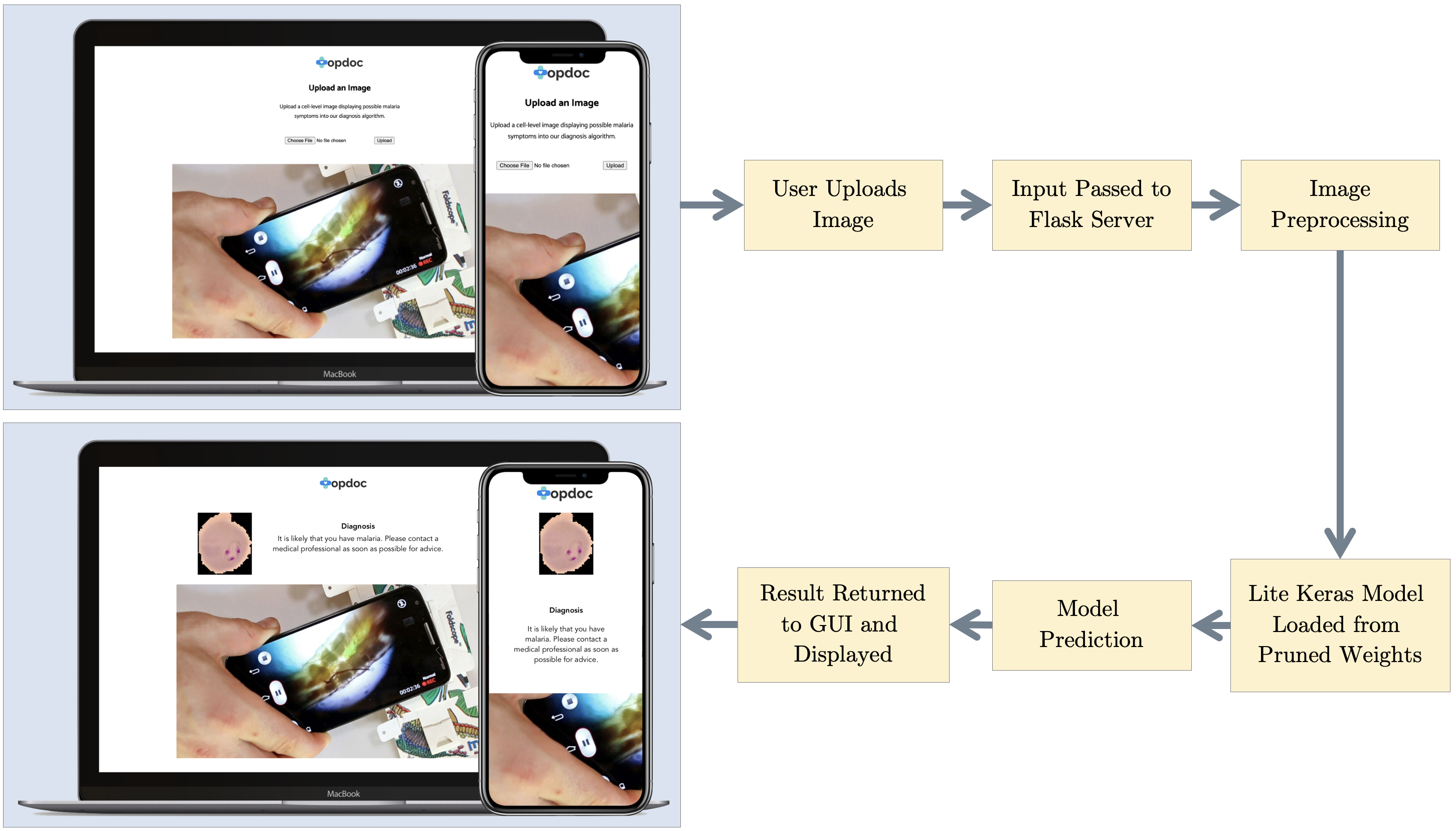}
\caption{Workflow of online malaria diagnosis tool.} \label{fig1}
\end{figure}

\subsection{CNN Architecture}

\begin{figure}
\includegraphics[width=\textwidth]{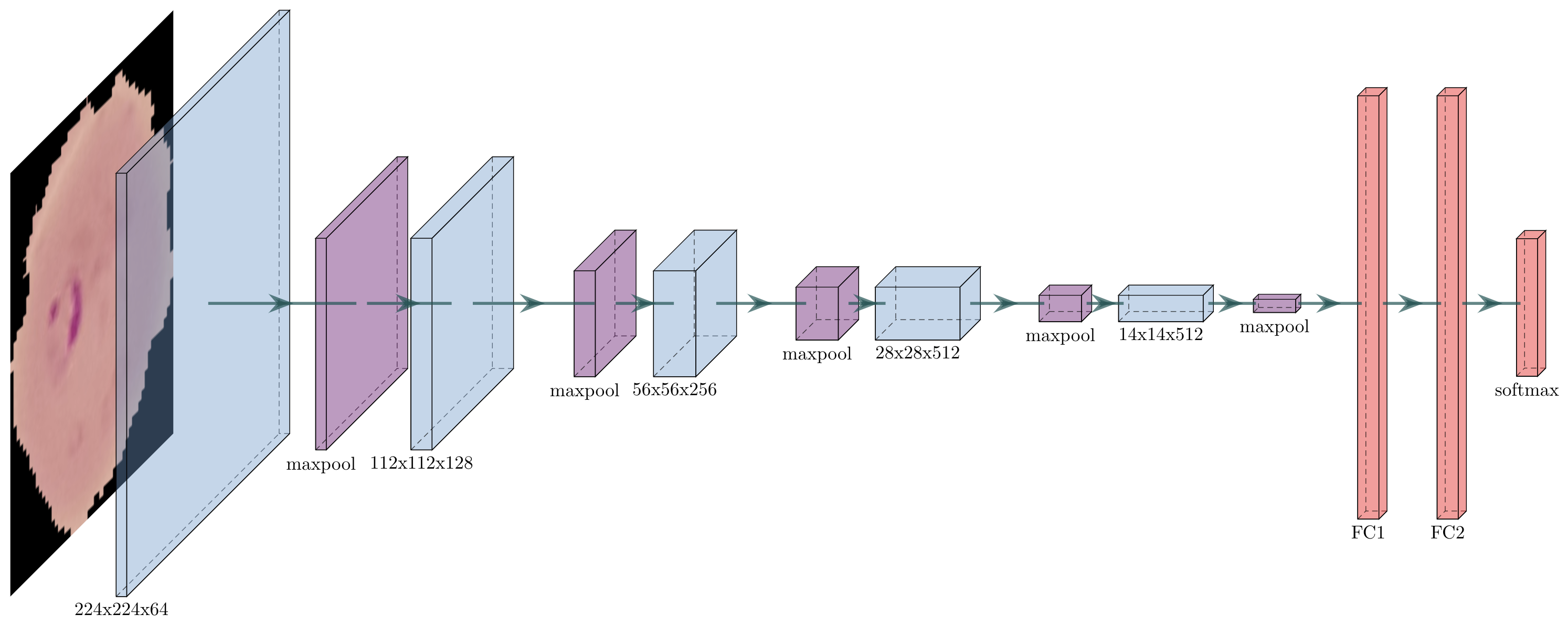}
\caption{VGG-19 model architecture.} \label{fig2}
\end{figure}

\noindent
Our CNN is based upon the concept of deep transfer learning—leveraging an existing model and applying its knowledge to a different context and dataset. First, we rebuilt the VGG-19 model (a 19-layer CNN as shown in Fig. 2) to perform feature extraction (a form of dimensionality reduction)  by using TensorFlow to add five dense layers to the end of the model [7]. This reduces the overfitting present in a basic malaria diagnosis CNN. Next, we used Keras to apply random transformations to the training images in order to augment the data and further reduce overfitting. Finally, we trained the model on the NIH dataset for 25 epochs.

\subsubsection{CNN Evaluation.}
Our CNN achieved the following results and outperformed state-of-the-art baselines:\\

\noindent
\begin{tabularx}{\textwidth}{|Y|Y|Y|}
\hline

  \textbf{Model} &  \textbf{Accuracy} & \textbf{F-score}\\
\hline
\textbf{Ours} &  \textbf{98.0\%} & \textbf{97.3\%} \\
\hline

Rajaraman et al. [8] &  95.9\% & N/A\\
\hline

Pattanaik et al. [9] &  N/A & 96.5\%\\
\hline

Masud et al. [10] &  97.3\% & 97.0\% \\
\hline
Shekar et al. [11] &  96.0\% & 96.0\% \\
\hline
\end{tabularx}\\\\

Moreover, our CNN is more accessible to end users. In the status quo, one of the most accessible malaria diagnosis algorithms is in the form of an Android app [12]; our algorithm can be easily used on any electronic device with internet access as shown in Fig. 1.

\subsection{DNN Architecture}
Our DNN architecture is adapted from Lin et al.'s work on learning efficient point cloud generation models for 3D object reconstruction [13]. Using a PyTorch-based procedural content generation (PCG) model similar to theirs (the model is founded upon an encoder-based structure generator that learns the general shape of an object based on a dataset of rendered depth images stored in an NPY file, a file format provided by the Python library NumPy), we convert a 2D image of a parasitic cell into a 3D model of that cell, stored as a Point Cloud Data (PCD) file. We then convert that PCD file into an OBJ file and load it into the echoAR API, retrieving the associated QR code that allows users to view the 3D model in AR [14]. Finally, we upload the OBJ file to Sketchfab in order to render it directly on our website and allow users to pan through it. 

We aim to improve current telemedicine practices and foster online patient education by assisting doctors in providing virtual diagnoses and treatment to a wider array of patients. By viewing 3D models of their patient’s cells in AR and also panning through those models via our website, doctors can better understand their patients’ conditions and more appropriately address them; they can then use those models to explain underlying pathology and necessary treatment to their patients. Through these educational tools, we also hope to improve patient health literacy, which has been shown to parameterize a patient’s adherence to treatment.

\section{Future Work}
\subsection{Limitations}
While there are numerous benefits to self-diagnosis assisted by electronic devices, there exist certain drawbacks. Patients who may be less inclined to use or are intimidated by electronic devices may not be able to properly engage in self-diagnosis and hence are at greater risk than technologically savvy patients [15]. Moreover, there exists the issue that not all patients in under-resourced areas or of low socioeconomic brackets can obtain an electronic device with internet access, rendering them unable to capitalize on our diagnostic and 3D modeling tools. To solve this problem, our CNN and DNN architectures can potentially be integrated into a low-cost handheld microscope such as the Foldscope, allowing patients to engage in end-to-end diagnosis with only a microscope rather than first obtaining an image using a microscope and then uploading that image to our website via an electronic device [16].

\subsection{Next Steps}
While our CNN is complete, we are currently working on improving our DNN and deploying it on our website in the same manner that our CNN has been deployed.

Our ultimate goal is to distribute our diagnostic and 3D modeling tools. We hope to build kits that can be distributed to individuals with limited access to medical facilities, including members of homeless respite centers or residents of rural, under-resourced regions where equipment such as microscopes is scarce. We hope to include vaculet blood collection sets for patients to obtain blood samples as well as low-cost handheld/smartphone microscopes such as the Foldscope, which our algorithms are compatible with.

Moreover, our diagnosis and 3D modeling solutions extend far beyond malaria detection. While our work is disruptive in fully automating the diagnosis of parasitic, cell-level diseases, our technologies can be applied to various other maladies. We believe that strengthening doctor-patient relationships—specifically in virtual spaces—and empowering patients to take charge of their own health is the future of medicine.

\section{Code Availability}
All of our code and model weights are open source and available on GitHub under the MIT License at \href{https://github.com/rvignav/E2EMD}{\texttt{https://github.com/rvignav/E2EMD}}.

\section{Acknowledgement}
The author would like to acknowledge Ayaan Haque for his contribution to the development of the malaria diagnosis CNN. He helped build the dense layers at the end of the model and took part in training the CNN on the NIH dataset.

\section{References}
\begin{enumerate}
    \item “Malaria in Africa.” UNICEF DATA, https://data.unicef.org/topic/child-health/malaria/. Accessed 6 July 2021.
\item Diagnostic Testing for Malaria. https://www.who.int/teams/control-of-negl\\ected-tropical-diseases/yaws/diagnosis-and-treatment/global-malaria-progra\\mme. Accessed 6 July 2021.
\item Prevention, CDC-Centers for Disease Control and. CDC - Malaria - Diagnosis \& Treatment (United States) - Diagnosis (U.S.). 31 Jan. 2019, https://www.\\cdc.gov/malaria/diagnosis\_treatment/diagnosis.html.
\item Cunningham, Jane, et al. “A Review of the WHO Malaria Rapid Diagnostic Test Product Testing Programme (2008–2018): Performance, Procurement and Policy.” Malaria Journal, vol. 18, no. 1, Dec. 2019, p. 387. BioMed Central, doi:10.1186/s12936-019-3028-z.
\item Naicker, Saraladevi, et al. “Shortage of Healthcare Workers in Developing Countries--Africa.” Ethnicity \& Disease, vol. 19, no. 1 Suppl 1, 2009, pp. S1-60–64.
\item Rajaraman, Sivaramakrishnan, et al. “Performance Evaluation of Deep Neural Ensembles toward Malaria Parasite Detection in Thin-Blood Smear Images.” PeerJ, vol. 7, May 2019, p. e6977. DOI.org (Crossref), doi:10.7717/peer\\j.6977.
\item Simonyan, Karen, and Andrew Zisserman. “Very Deep Convolutional Networks for Large-Scale Image Recognition.” ArXiv:1409.1556 [Cs], 6, Apr. 2015. arXiv.org, http://arxiv.org/abs/1409.1556.  
\item Rajaraman, Sivaramakrishnan, et al. “Pre-Trained Convolutional Neural Networks as Feature Extractors toward Improved Malaria Parasite Detection in Thin Blood Smear Images.” PeerJ, vol. 6, 2018, p. e4568. PubMed, doi:10.7717/peerj.4568.
\item Pattanaik, Priyadarshini Adyasha, et al. “Deep CNN Frameworks Comparison for Malaria Diagnosis.” ArXiv:1909.02829 [Cs, Eess], Sept. 2019. arXiv.org, http://arxiv.org/abs/1909.02829.
\item Masud, Mehedi, et al. “Leveraging Deep Learning Techniques for Malaria Parasite Detection Using Mobile Application.” Wireless Communications and Mobile Computing, vol. 2020, July 2020, p. e8895429. www.hindawi.com, doi:10.1155/2020/8895429.
\item Shekar, Gautham, et al. “Malaria Detection Using Deep Learning.” 2020 4th International Conference on Trends in Electronics and Informatics (ICOEI)\\(48184), 2020, pp. 746–50. IEEE Xplore, doi:10.1109/ICOEI48184.2020.9143\\023.

\item Yang, Feng, et al. “Smartphone-Supported Malaria Diagnosis Based on Deep Learning.” Machine Learning in Medical Imaging, edited by Heung-Il Suk et al., Springer International Publishing, 2019, pp. 73–80. Springer Link, doi:10.1007/978-3-030-32692-0\_9.
\item Lin, Chen-Hsuan, et al. “Learning Efficient Point Cloud Generation for Dense 3D Object Reconstruction.” ArXiv:1706.07036 [Cs], June 2017. arXiv.\\org, http://arxiv.org/abs/1706.07036.
\item EchoAR Queries. https://docs.echoar.xyz/queries. Accessed 6 July 2021.
\item Ozdalga, Errol, et al. “The Smartphone in Medicine: A Review of Current and Potential Use Among Physicians and Students.” Journal of Medical Internet Research, vol. 14, no. 5, Sept. 2012, p. e128. PubMed Central, doi:10.2196/jmir.1994.
\item “Foldscope Instruments $|$ Magnify Your Curiosity.” Foldscope Instruments, https://www.foldscope.com. Accessed 6 July 2021.

\end{enumerate}

\end{document}